%% file: root1.tex
\documentclass[letterpaper, 10 pt, conference]{ieeeconf}  

\IEEEoverridecommandlockouts                              

\usepackage{amsmath,amssymb,amsfonts}
\usepackage{algorithmic}
\usepackage{graphicx}
\usepackage{textcomp}
\usepackage{xcolor}
\usepackage{subfigure} 

\newtheorem{theorem}{Theorem}
\newtheorem{lemma}{Lemma}
\newtheorem{remark}{Remark}
\newtheorem{definition}{Definition}

\input{newcommands.tex}

\usepackage{algorithm}
\usepackage{algorithmic}

\newcommand{\x}{\hat{x}_u^t}

\title{\LARGE \bf
Online Scoring with Delayed Information: A Convex Optimization Viewpoint
}

\author{Avishek Ghosh \\
Department of EECS\\
UC Berkeley, CA USA\\
avishek\_ghosh@berkely.edu\\
\and
Kannan Ramchandran \\
Department of EECS\\
UC Berkeley, CA USA\\
kannanr@eecs.berkeley.edu
}

\begin{document}

\maketitle
\thispagestyle{empty}
\pagestyle{empty}

\begin{abstract}

We consider a system where agents enter in an online fashion and are evaluated  based on their attributes or context vectors. There can be practical situations where this context is partially observed, and the unobserved part comes after some delay. We assume that an agent, once left, cannot re-enter the system. Therefore, the job of the system is to provide an estimated score for the agent based on her instantaneous score and possibly some inference of the instantaneous score over the delayed score. In this paper, we estimate the delayed context via an online convex game between the agent and the system. We argue that the error in the score estimate accumulated over $T$ iterations is small if the regret of the online convex game is small. Further, we leverage side information about the delayed context in the form of a correlation function with the known context. We consider the settings where the delay is fixed or  arbitrarily chosen by an adversary. Furthermore, we extend the formulation to the  setting where the contexts are drawn from some Banach space. Overall, we show that the average penalty for not knowing the delayed context while making a decision scales with $\mathcal{O}(\frac{1}{\sqrt{T}})$, where this can be improved to $\mathcal{O}(\frac{\log T}{T})$ under special setting.
\end{abstract}

\section{Introduction}
Our problem is motivated by the following scenarios: for example, typically tech companies administer a test to evaluate if candidates are suitable for the job. Also, we can assume the candidates to be active job-seekers and are applying to several other companies. Thus, it is of the companies interest to complete the grading process as quickly as possible, since delays may
result in a qualified candidate accepting a job at some another company.
The test consists of two different types of questions: the first type are those that can be automatically graded (e.g. multiple choice questions) and thus the scores for those are obtained instantaneously, and questions of the second type are need to be human-graded as they are more involved (e.g. subjective type questions). Assuming that the human grading takes a non-trivial amount of time,  if the company waits for the human graded scores before informing candidates, they may miss out on hiring qualified candidates. Thus there is a strong incentive for the company to provide an estimated score based on just the performance on the automatically graded questions. The estimated score allows the candidate to infer their rank. Of course, we want the estimated score to be close to the true score based on the both the automatically graded and human graded scores. 

Another setting where our formulation is relevant is in the context of generating crowd-sourced questions (e.g., for massive online open courses). Candidates take a test in which they answer a set of questions (which are automatically graded), but they are also asked to create one or more new questions for future candidates to answer (which have to be evaluated manually). The test is scored based on their performance on questions they answer and the quality of those that they create, but the candidate is given an estimate of their score before the quality of their questions is evaluated.  We see that this and the previous problem are special cases of the following abstract problem:

\textit{
\paragraph*{Problem} In an online setting where a system outputs scores based on  context, that are part-observed and part-delayed, can we estimate the scores efficiently such that they are close to the true score (with full context)? }

\subsection{Related work and Our Contribution}

There has been a series of work in the online learning literature (\cite{kalai2005efficient}, \cite{shalev2012online}, \cite{Blum:1998} to name a few) where a system learns the environment in an adaptive fashion based on the feedback it receives from the agents. These ideas can be generalized to the setting where the system has partial information about the agent. There are several ways to deal with this setting, for example, classically one assumes the delayed data as ``missing data'' and perform data imputation. If the distribution from which the data is generated is known, a na\"ive approach is ``mean-imputation'', i.e., replace the missing data by the mean. This could work if the distribution of interest is sub-gaussian or more generally sub-exponential where the data points concentrate around the true mean with high probability. In \cite{baraldi_10,van_12} the authors explore a generalized version of problem-dependent imputation.

If we have reasons to believe that there is some correlation between the observed and the unobserved context, another systematic approach is to estimate the unobserved part from the observed context via a correlation function (\cite{joulani_13,mesterharm_05}). Note that, our situation is a little delicate, where the information is not lost but merely delayed. So one can leverage the techniques used in delayed online learning in our scenario. Indeed, \cite{langford_09}, \cite{quanrud_15}, and \cite{Joulani_16} provide a convex optimization formulation for the delayed data as a function of delay. In this paper, we intend to combine the above-mentioned two approaches.

 In this paper, we do not assume that the contexts are coming from a known distribution. Still we can use techniques from delayed online optimization literature (which typically have no distributional assumption). Also, we do not assume that the unknown context is completely determined via a correlation function of the observed context, rather we allow the estimate of the unknown context to be partially influenced by the observed context. We develop several online learning algorithms that capture both delay in information and correlation from the observed context. Our contribution can be summarized as follows:
 
\begin{itemize}

    \item We provide an online convex optimization formulation for the estimation of the delayed data, and also allow a correlation function to influence the estimation process. We show that, the estimation error is directly related to the cumulative regret of the online convex optimization algorithms, and so we focus on minimizing cumulative regret under different loss function scenarios, e.g., convex, strongly convex etc.
    
    \item We present novel delayed online gradient descent and online mirror descent algorithm with a correlation function, and provide theoretical analysis for cumulative regret.
    
    \item We analyze the delayed and correlated online convex programs both with fixed delay, and adversarially chosen (arbitrary) delay and prove theoretical upper bounds for regret as a function of delay.
    
    \item We verify our theoretical findings via simulations. We compare the performance of our algorithms with a sample mean based naive heuristic algorithm. We observe that, when data is not coming from a well behaved distribution (for example, we let data points drawn randomly from a pentagon), the naive algorithm performs very poorly in comparison to our newly developed algorithms.
\end{itemize}

 \vspace{-2mm}
 \subsection{System Model}
\label{sec:system_model}
 We now present the system model that will be used throughout the paper. At time $t$ assume an agent with context $x^t \in \mathbb{R}^d$ enter the system. A part of $x^t$, $x_k^t \in \mathbb{R}^{d_1}$ is revealed immediately, where the rest, $x_u^t \in \mathbb{R}^{d_2}$ is unobserved and revealed after some delay. Hence, $x^t= (x_k^t,x_u^t)$, and we assume $d_1 \geq d_2$ (note that this is not a strict requirement, we assume this to get a canonical representation of the influence function $\Phi(.)$, see Section~\ref{sec:fixed_delay}). The agent is given a score by the system based on $x^t$, and we assume that there exists a scoring function $g(x^t)= g(x_k^t,x_u^t)$. An example would be: $g(x^t)=c_1 x_k^t + c_2 x_u^t$, where $ c_1$, and $c_2 $ are positive constants. The job of the system is to output an estimate of $x_u^t$, $\hat{x}_u^t$, such that the estimated score, $g(x_k^t, \hat{x}_u^t)$ is close to the true score $g(x_k^t,x_u^t)$. Note that there is no distributional assumptions on $x_k^t$ and $x_u^t$. Rather we let $x_k^t$ and $x_u^t$ be sampled from an arbitrary convex set $\mathcal{K}$. We also assume that $\hat{x}_u^t$ is influenced by $x_k^t$ via an influence (correlation) function $\Phi : \mathbb{R}^{d_1} \rightarrow \mathbb{R}^{d_2}$.

 We intend to obtain $\hat{x}_u^t$ via online convex programming. We model the scenario as a convex game between agents and the system. At each round $t$, an agent chooses $\hat{x}_u^t$ from a convex body $\mathcal{K}$. Simultaneously, without observing $\hat{x}_u^t$, the system chooses a convex function $f_t$ and the agent incurs a loss of $f_t(\x)$. Since, the objective of the system is to estimate $x_u^t$, the loss functions $f_t(\x)$ are constrained to have additional structures. Here we restrict that the loss functions should be of the form, $f_t(\x)=f(\twonms {\x-x_u^t})$, where  $f:\R^+ \cup \lbrace0 \rbrace \rightarrow \R^+ \cup \lbrace0 \rbrace $ satisfies: (a) $f(u)$ is an increasing function of $u$ and (b) $f(0)=0$. We denote the class of convex functions satisfying the mentioned conditions by $\mathcal{F}$, and hence, $f_t \in \mathcal{F}$. There can be several examples of convex functions belonging to this class:

\begin{itemize}
\item $f_t(\x)=f(\twonms {\x-x_u^t}) = \twonms{\x-x_u^t}$.
\item $f_t(\x) = (\x-x_u^t)^T(\x-x_u^t)=\twonms{\x-x_u^t}^2$ i.e., (quadratic loss). This can be generalized to, $ f_t(\x)= \twonms{\x-x_u^t}^{m}$, where $m \in \mathbb{Z}_+$.
\item $f_t(\x)= a\exp \bigg (\frac{\twonms{\x-x_u^t}^{m}}{\sigma_1^2} \bigg )$, for some constants $a, \sigma_1$ and $m\in \mathbb{Z}_+$.
\end{itemize} 
    
  The goal of online convex programming is to choose $\x$ sequentially such that the total loss, defined over the sum of losses over $T$ episodes is minimized.  The online convex program ensures that the total loss, defined over the sum of losses over $T$ episodes of the play is small (refer to \cite{hazan_oco} for details), i.e., $ \sum_{t=1}^{T} f_t(\x) \leq \min_{y \in \mathcal{K}} \sum_{t=1}^{T} f_t (y) + r(T)$, where a typical example would be $r(T)= \mathcal{O}(\sqrt{T})$ and this is achieved by several optimization algorithms including online gradient descent (\cite{zinkevich2003online}). Notice that, in the upper bound on total loss,  $\min_{y \in \mathcal{K}} \sum_{t=1}^{T} f_t (y)$ is the loss in the best case scenario, where the agent gets to see all the loss functions and gets to select a fixed decision that minimizes the sum loss over $T$ episodes. As seen in the online convex optimization literature (\cite{hazan_oco},\cite{zinkevich2003online}), this term is unavoidable, and hence the performance of an algorithm is measured relative to this fixed decision optimal loss. From now on, we denote, $D^* := \min_{y \in \mathcal{K}} \sum_{t=1}^{T} f_t (y)$.

 We further assume that the scoring function is separable, i.e., there exist functions $g_1: \R^{d_1} \rightarrow \R^+ \lbrace0 \rbrace$ and $g_2: \R^{d_2} \rightarrow \R^+ \lbrace0 \rbrace$, such that, $g (x_k^t,x_u^t) = g_1(x_k^t)+g_2(x_u^t)$. A simple example would be: $g_1(t)=c_1 x_k^t$, $g_2(t)=c_2 x_u^t$ for some $c_1>0,\,\, c_2 >0$. Also, we assume $g_2$ to be $1$-lipschitz. Under these assumptions, the estimation error in scores under the loss function $f$ and with $T$ episodes will be,
 \vspace{-2mm}
\begin{eqnarray}
&&\sum_{t=1}^{T}f\bigg ( |g(x_k^t,\x)-g(x_k^t,x_u^t)| \bigg ) \nonumber \\
 &= & \sum_{t=1}^{T}f\bigg ( | g_1(x_k^t)+g_2(\x) - g_1(x_k^t)-g_2(x_u^t)| \bigg ) \nonumber \\
&\leq & \sum_{t=1}^{T}f \bigg (  \twonms { \x - x_u^t} \bigg ) 
=  \sum_{t=1}^{T} f_t(\x) \leq  D^* + r(T) \nonumber
\end{eqnarray}

where we use the Lipschitz property of $g_2$, and the properties of the loss function $f$. So, we can see that if we can control $r(T)$, we have a control over the cumulative error in estimation. In the following sections, we propose several iterative algorithms to obtain $\x$ such that $r(T)=o(T)$, i.e., sublinear in $T$.

\section{Score Estimation using Online Convex Optimization With Fixed Delay}
\label{sec:fixed_delay}

 We now consider the case where delay in the unknown context, $x_u^t$ is fixed and known to the system. We denote the delay by $\tau$. We will analyze $3$ different cases under fixed delay: (1) when the loss function, $f_t(.)$ is convex, (2) when $f_t(.) $  is strongly convex and (3) when we relax the condition that $x_k^t$ and $x_u^t$ belong to subsets of $\R^{d_1}$ and $\R^{d_2}$ respectively; rather assuming that $x_k^t$ and $x_u^t$ belong to some generalized Banach space and use mirror-descent type algorithm (\cite{hazan_oco}). All the proofs are deferred to Appendix~\ref{appendix:fixed_delay}.
 
 \subsection{Online learning of convex loss function under fixed delay}
 \label{subsec:convex_fixed}
 
 We proceed to minimize the total loss accumulated over $T$ rounds, under the assumption that the delay in observing the loss function is fixed over all rounds. We stick to the System model of \ref{sec:system_model}. Let $\tau \in \mathbb{N}$ be the fixed delay. We also assume a correlation function, $\Phi$ capturing the influence of $x_k^t$ over $x_u^t$. The details are formally stated in Algorithm~\ref{algo:convex_fixed}.
 
\begin{algorithm}[!hbtp]
	\caption{Delayed Correlated Online Gradient Descent} \label{algo:convex_fixed}
	\begin{algorithmic}[1]	
	\STATE Input: Step size $\eta_t$, delay $\tau \in \mathbb{N}$, influence function $\Phi_{t}(.)$ with weight $\beta_t$.

	\STATE Initialization: set $\hat{x}_u^1,\ldots,\hat{x}_u^\tau =0$

	\STATE For iteration $t=\tau+1$ to $T$ do \\
	\begin{itemize}
	\item Obtain $f_t$, incur loss: $f_t(\x)$
	\item Compute Gradient of the latest completely known loss: $g_{t-\tau}=\nabla f_{t-\tau}(\hat{x}_u^{t-\tau})$
	\item Update: $\hat{x}_u^{t+1}= \x - \eta_t g_{t-\tau} + \beta_t \Phi_{t+1}(x_k^{t+1})$ 
	\end{itemize}
    
	\end{algorithmic} 
\end{algorithm}

 If at any time step, $\hat{x}^{t+1}_u$ lies outside the convex set $\mathcal{K}$, we project it back to $\mathcal{K}$ via an euclidean projection. Since $\mathcal{K}$ is convex, euclidean projection exists, and is unique and $1$-Lipschitz (contraction), it is sufficient to analyze the scenario without projection \cite{zinkevich2003online}.

  At time $t+1$, the most recent completely known function is $f_{t-\tau}$, since the information about the loss function is delayed by $\tau$. For, $\tau = 0$, according to the format of the convex game, the most recent completely known function would be $f_t$. The algorithm chooses a direction which is a combination of a greedy direction where the most recent loss function is minimized and a direction provided by the influence function.

   Also, since the delay is $\tau$, we set the first $\tau$ decisions to $0$. In the online scoring of agents scenario, we can think that, in the first $m \tau$ iterations of the algorithm (where, $m \in \mathbb{Z}_+$, a small positive number, $2$ for example), it provides scores for dummy candidates, before the real agents are being scored. This is to smooth out the crude initialization of $\x$ for $1 \leq t \leq \tau$. If we run the algorithm for $T$ rounds, the regret with respect to the fixed best decision is defined as,
\begin{equation}
R(T)=\sum_{t=1}^{T} f_t(\x)- \min_{x \in \mathcal{K}} f_t (x) \nonumber
\end{equation}
 
We will prove an upper bound on $R(T)$ as a function of $\tau, T$ for a particular characterization of influence function $\Phi(.)$, via a number of steps, borrowing a few techniques from \cite{langford_09}.

Let $x^* = \arg \min_{x \in \mathcal{K}} \sum_{t=1}^{T} f_t(x)$. Since the functions $f_t$ are convex, $R(T)$ satisfies,
\begin{equation}
R(T) \leq \sum_{t=1}^{T} \langle \nabla f_t(\x),\x-x^* \rangle= \sum_{t=1}^{T} \langle g_t, \x-x^* \rangle \nonumber
\end{equation}

Define, $\tilde{g}_{t-\tau} = g_{t-\tau} -\frac{\beta_t}{\eta_t}\Phi_{t+1}(x_k^{t+1})$, and so,
\begin{equation}
R(T) \leq \sum_{t=1}^{T}  ( \langle \tilde{g}_t, \x-x^* \rangle + \langle \frac{\beta_{t+\tau}}{\eta_{t+\tau}}\Phi_{t+\tau+1}(x_k^{t+\tau+1}) ,\x-x^* \rangle ) \nonumber
\end{equation}

We now define a deviation function, $D(x_1||x_2)=\frac{1}{2} \norms{x_1-x_2}^2$ for all $x_1, x_2 \in \mathcal{K}$. We have the following result:

\vspace{2mm}
\begin{lemma}
\label{lem:divergence}
For all $t > \tau$, we have:
\begin{eqnarray}
&&\langle \hat{x}^{t-\tau}_u -x^*, \tilde{g}_{t-\tau} \rangle \leq  \frac{1}{2}\eta_t \norms{\tilde{g}_{t-\tau}}^2+\frac{D(x^*||\x)}{\eta_t} \nonumber \\
&& - \frac{D(x^*||\hat{x}^{t+1}_u)}{\eta_t}+ \sum_{k=1}^{\min(\tau,t-(\tau+1))} \eta_{t-k} \langle \tilde{g}_{t-\tau-k},  \tilde{g}_{t-\tau} \rangle \nonumber
\end{eqnarray}
\end{lemma}

\paragraph*{Assumptions}
We assume that the loss functions are $L$ lipschitz ($L >0$). Since the functions $f_t$ are convex, this implies, $\norm{\nabla f_t(x)} \leq L$.
We also assume a canonical form of the influence function $\Phi(.)$. For simplicity, from now on we assume, $d_1=d_2$ (if $d_1 \geq d_2$, one can project $x_k^{t+1}$ on a $d_2$ dimensional space and use the projected vector). We take, $\Phi_{t+1}(x_k^{t+1})=\lambda_{t+1} x_k^{t+1}$. If we have reasons to believe that the correlation coefficient between $x_k^{t+1}$ and $x_u^{t+1}$ is positive, we take $\lambda_{t+1} > 0$, otherwise $\lambda_{t+1} < 0$. We choose the parameters of the algorithm as follows: $\eta_t = \beta_t = |\lambda_{t+1}| = \frac{\sigma}{\sqrt{t-\tau}}$, for $t>\tau$ and $0$ otherwise, with some $\sigma >0$. Furthermore, $\norms{\x} \leq R$, and $\norms{x_k^t} \leq R$ for all $t$, and so, $D(x_1||x_2) \leq 2R^2$ for any $x_1,x_2 \in \mathcal{K}$.

With the above assumptions, we have
\begin{eqnarray}
\norms{\tilde{g}_{t-\tau}} &\leq & \norms{g_{t-\tau}} + \frac{\beta_t}{\eta_t}|\lambda_{t+1}| \norms{x_k^{t+1}} \leq L + \frac{\sigma}{\sqrt{t-\tau}}R \nonumber \\
 &\leq & L + \sigma R:=L' \,\,\, \mbox{for} \,\,\, t-\tau \geq  1 \nonumber
\end{eqnarray}

\begin{theorem}
\label{thm:convex_fixed}
With the given choices of $\eta_t$, $\beta_t, |\lambda_{t+1}|$ and $\sigma:=\frac{R}{L'\sqrt{\tau}}$, the cumulative Regret of the Delayed Correlated Online Gradient Descent is given by,
\begin{equation}
R(T) \leq  C_1 \sqrt{\tau T} \nonumber
\end{equation}
where $C_1$ is a constant independent of $T$ and $\tau$.
\end{theorem}

\subsection{Online Learning of Strongly Convex loss function}
\label{subsec:strongly_convex_fixed}

We now analyze the setting where the loss functions, $f_t$ are strongly convex, with parameter $\gamma \,\,(>0)$. We choose the step-size $\eta_t = \frac{1}{\gamma (t-\tau)}$ for all $t>\tau$ and $0$ otherwise. We keep the other assumptions identical to the previous section. We have the following bound on the regret of the correlated delayed online gradient descent algorithm.

\begin{theorem}
\label{thm:strongly_convex_fixed}
With $\eta_t = \beta_t = |\lambda_t| = \frac{1}{\gamma (t-\tau)}$ (for $t > \tau$), the regret of Algorithm~\ref{algo:convex_fixed}, for $\gamma$ strongly convex loss,
\small
\begin{equation}
R(T) \leq 2\gamma \tau R^2+ \frac{2R^2}{\gamma} H(T) + (\frac{1}{2}+\tau) L'^2 \frac{1}{\gamma} H(T-\tau) \leq C_2 \tau \log T \nonumber
\end{equation}
where, $H(.)$ denotes harmonic number, $C_2$ is a constant, independent of $T$ and $\tau$.
\end{theorem}

\begin{remark}
The regret scaling with respect to $T$ is better in this case, since we leverage the fact that the loss functions are strongly convex. However, the scaling with respect to $\tau$ is worse in this situation. It is linear in this case, in contrast to $\sqrt{\tau}$ for the weakly convex losses.
\end{remark}

\subsection{Online Correlated Delayed Mirror Descent Optimization for Generic Banach Spaces}

In this section, we generalize Algorithm~\ref{algo:convex_fixed} to a setting, where, $\x \in \mathbb{B}^{d_2}$, and $x_k^{t} \in \mathbb{B}^{d_1}$, where $\mathbb{B}$ is a Banach space (a complete normed space with norm $\norms{.}$) Like the previous section, we assume $d_1=d_2$. Correspondingly, define a mirror map, $M:\mathbb{B}^{d_1} \rightarrow \R$. Then, the Bregmen divergence (\cite{hazan_oco}, Chapter 5) between $x$ and $y$ ($\in \mathbb{B}^{d_1}$), with mirror map $M$ is given by,
$$
D_M (x||y) =M(x)-M(y) - \langle x-y, \nabla M(y) \rangle
$$
Also, a loss function $f_t$ is said to be $\gamma$ strongly convex with respect to map $M$, if, for $x,y \in \mathbb{H}^{d_1}$
$$
f(x) - f(y) - \langle x-y, \nabla f(y) \rangle \geq \gamma D_M(x||y)
$$

Furthermore, given a convex function $f$, the Fenchel dual of $f$ is defined as,
$f^*(y) = \sup_x \langle x,y \rangle -f(x)$. Using these set of definitions, we now present the generalized version of Algorithm~\ref{algo:convex_fixed}:

\begin{algorithm}[!hbtp]
	\caption{Delayed Correlated Online Mirror Descent} \label{algo:convex_fixed_mirror}
	\begin{algorithmic}[1]	
	\STATE Input: Mirror Map $M$, step size $\eta_t$, delay $\tau \in \mathbb{N}$, influence function $\Phi_{t}(.)$ with weight $\beta_t$.

	\STATE Initialization: set $\hat{x}_u^1,\ldots,\hat{x}_u^\tau =0$

	\STATE For iteration $t=\tau+1$ to $T+\tau $ do \\
	\begin{itemize}
	\item Obtain $f_t$, incur loss: $f_t(\x)$
	\item Compute Gradient of the last completely known loss: $g_{t-\tau}=\nabla f_{t-\tau}(\hat{x}_u^{t-\tau})$
	\item Update: $\hat{x}_u^{t+1}= \nabla M^* ( \nabla M( \x)  - \eta_t g_{t-\tau} + \beta_t \Phi_{t+1}(x_k^{t+1}) )$ 
	\end{itemize}
    
	\end{algorithmic} 
\end{algorithm}

Note that, if the mirror map, $M(x) = \frac{1}{2} \norms{x}^2$, then, $\nabla M(x) =x$, and $M=M^*$, so we get back Delayed Correlated Online Gradient Descent (Algorithm~\ref{algo:convex_fixed}). We will analyze the regret performance of this algorithm.

\begin{lemma}
If $M$ is strongly convex with respect to the norm corresponding to $\mathbb{B}^{d_1}$, then, the loss minimizer $x^*$ satisfies,
$$
\langle \x -x^*, \tilde{g}_{t-\tau} \rangle \leq \frac{D_M(x^*||\hat{x}_u^t)-D_M(x^*||\hat{x}_u^{t+1})}{\eta_t} + \frac{\eta_t}{2}\norms{\tilde{g}_{t-\tau}}^2_*
$$
where $\norms{.}_*$ is the dual norm of the norm associated with $\mathbb{B}^{d_1}$.
\label{lem:mirror}
\end{lemma}

This is a direct consequence of \cite{shalev2007logarithmic}, and hence we omit the proof.

We  assume, $\norms{\tilde{g}_{t-\tau}}_* \leq L'$,  $\norms{x_k^{t}}_* \leq R$ and $ \norms{\x}_* \leq R$, for all $t \geq 0$ . Note that, if we are working with $\ell_2$ norm, the dual norm is also $\ell_2$, and hence the assumptions are identical to that of Section~\ref{subsec:convex_fixed}. Also assume $d_1 =d_2$, and $\Phi(x_k^{t+1}) = \beta_t x_k^{t+1}$.

\begin{theorem}
\label{thm:mirror_fixed}
Suppose the mirror map, $M$ satisfies,
$$
\norms{ \nabla M^*(\nabla M(x) - y)-x} \leq L_M \norms{y}
$$
with $L_M >0$. With $\eta_t = \beta_t = \frac{\sigma}{\sqrt{t-\tau}}$, $\sigma^2 =\frac{R^2}{\tau L_M L'^2}$, the regret of Algorithm~\ref{algo:convex_fixed_mirror}, is given by,
$$
R(T) \leq C_3 \sqrt{L_M \tau T}
$$
where $C_3$ is a constant independent of $\tau$ and $T$.
\end{theorem}

\section{Formulation with Adversarial Delay}
\label{sec:adv_delay}
In this section, we assume that, the delays are not fixed; rather chosen in an adversarial way, with no assumptions or restrictions made on the action of the adversary. We start with some notation:

We continue to assume that each agent enters the system in an online fashion, and the job of the system is to output an estimated score based on partially observed context. For each $t >0$, let $\tau_t$ be some non-negative integer denoting delay, and $d_t =\tau_t +1$. The feedback from round $t$, is available, at the end of iteration $t+d_t-1$ and may be used by the system at round $t+d_t$. In the setting of Section~\ref{sec:fixed_delay}, $d_t = \tau_t +1=\tau +1$, and in the no-delay setting, $d_t = 1$. Let $\mathcal{F}_t = \lbrace u \in \lbrace 1,2,\ldots, T \rbrace : u+d_u -1 =t \rbrace$ be the index of the rounds whose feedback is available at round $t$. Also let $D=\sum_{i=1}^{T} d_t$, be the sum of all delays. We have, $\mathcal{F}_t = \lbrace t-\tau \rbrace$ and $D=\tau(T+1)$ for Section~\ref{sec:fixed_delay}. Similarly, $\mathcal{F}_t = \lbrace t  \rbrace$ and $D=T$ with $\tau_t=\tau =0$. The algorithm under this setting is given in Algorithm~\ref{algo:convex_adv_delay}.

\begin{algorithm}[!hbtp]
	\caption{Correlated Online Gradient Descent with Adversarial Delay} \label{algo:convex_adv_delay}
	\begin{algorithmic}[1]	
	\STATE Input: Step size $\eta$, set $\mathcal{F}_t$, influence function $\Phi_{t}(.)$ with weight $\beta $.

	\STATE Initialization: set $\hat{x}_u^1,\ldots,\hat{x}_u^\tau =0$

	\STATE For iteration $t=\tau+1$ to $T$ do \\
	\begin{itemize}
	\item Obtain $f_t$, incur loss: $f_t(\x)$
	\item Construct $\mathcal{F}_t$, and let $g_s = \nabla f_s (\hat{x}_u^s)$ 
	\item Update: $\hat{x}_u^{t+1}= \x - \eta \sum_{s \in \mathcal{F}_t} g_s  +  \beta \Phi_{t+1}(x_k^{t+1})$ 
	\end{itemize}
    
	\end{algorithmic} 
\end{algorithm}

 Since we are assuming that, $\x \in \mathcal{K}$, if the obtained $\hat{x}_u^{t+1}$ falls outside $\mathcal{K}$, we simply project it (via eucledian projection) back to $\mathcal{K}$. Since, $\mathcal{K}$ is convex, the eucledian projection is be a contraction ($1$ Lipschitz), and hence we are simply omitting the projection step in Algorithm~\ref{algo:convex_adv_delay}. Assuming $d_1 =d_2$ and $\Phi_t ( x_k^{t+1}) = \lambda x_k^{t+1}$, we now characterize the regret,

\begin{theorem}
\label{thm:convex_adversary}
Let $\eta = \beta$ and $\lambda$ be fixed. In presence of adversarial delay, the regret of Algorithm~\ref{algo:convex_adv_delay} is given by,
$$
R(T) \leq C_4  (\frac{1}{\eta}+ \eta C_5 (T + D) ) \leq C_6 \sqrt{D}
$$
when the last inequality holds if we choose $\eta = \mathcal{O}(1/\sqrt{T+D})$. $C_4,C_5$ and $C_6$ are constants.
\end{theorem}

\section{Simulations}
\label{sec:simulations}

\begin{figure}[t!]
    \centering
    \includegraphics[height = 0.25\textwidth, width=0.4\textwidth]{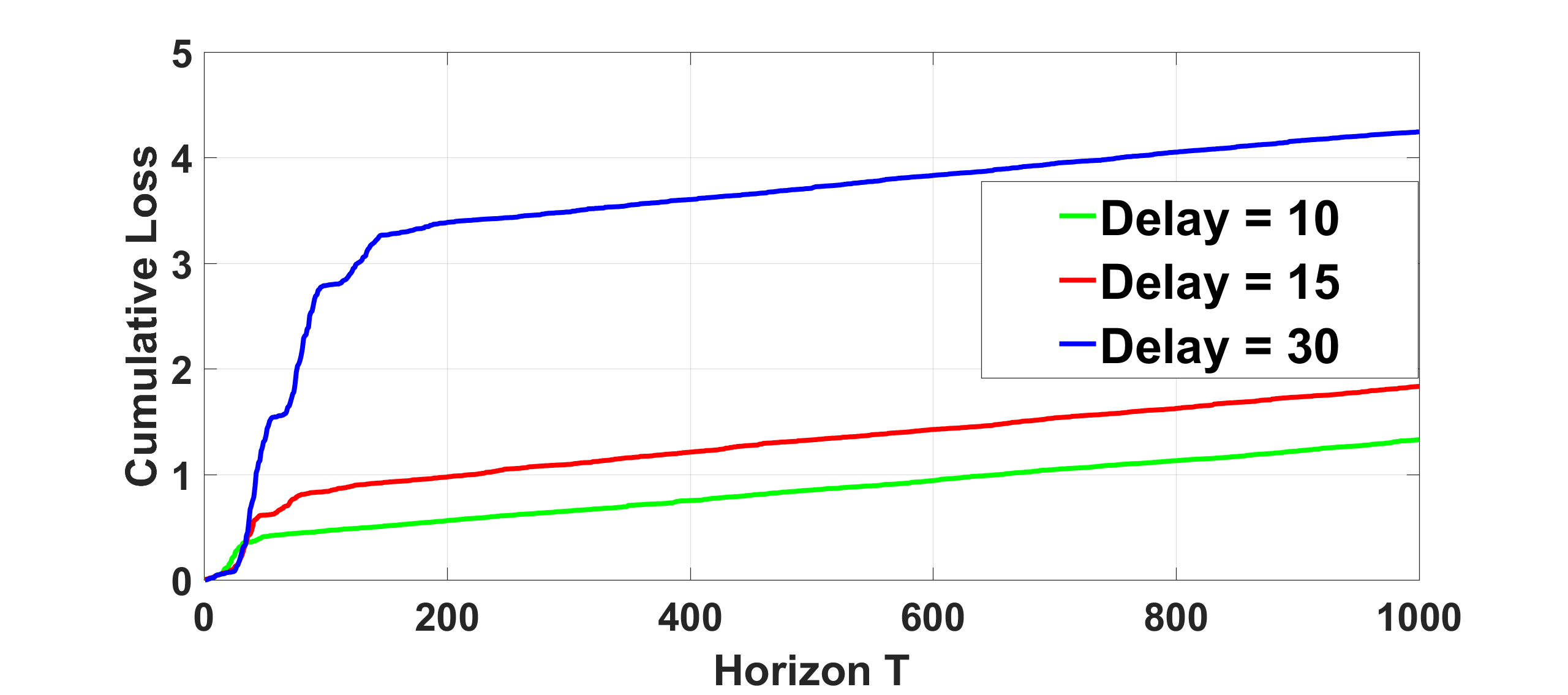}
    \vspace{-1mm}
    \caption{The evolution of cumulative loss with respect to time for different delays. Each point in the plot is an average over $100$ trials.
     }
    \label{fig:delay}
\end{figure}

In this section we present empirical validations of the results presented in the previous sections. First we will demonstrate how the performance of the system changes with different delay. We stick to the framework of Section~\ref{sec:system_model} and take $d_1=d_2=1$. We generate correlated gaussian random variables (for $x_u^t$ and $x_k^t$) with unit mean and variance and choose the following loss function: $f_t(\x)= a \norms{\x-x_u^t}^2 + b $ (quadratic loss), where $a$ and $b$ are drawn uniformly at random from $[0,1]$ at each iteration by the adversary. With a correlation factor of $0.5$ and the step size of $\frac{0.5}{\sqrt{t-\tau}}$, where $\tau$ is the delay, we run the correlated delayed online gradient descent algorithm with $\tau = 10 , 15$ and $30$, and compute the cumulative loss function (i.e., sum of the loss function) over the number of iterations (i.e., horizon). The result is shown in Figure~\ref{fig:delay}. Firstly, the cumulative loss behaves in a logarithmic fashion with respect to horizon, which is predicted by the theory. Also, the loss increases as $\tau$ increases. Hence the performance degrades as the information is further delayed. Also, Section~\ref{subsec:strongly_convex_fixed} suggests that the cumulative loss upper bound should behave linearly with $\tau$. From Figure~\ref{fig:delay}, we observe that this holds true for higher values of horizon $T$.

\begin{figure}[t!]
    \centering
    \includegraphics[height = 0.25\textwidth, width=0.4\textwidth]{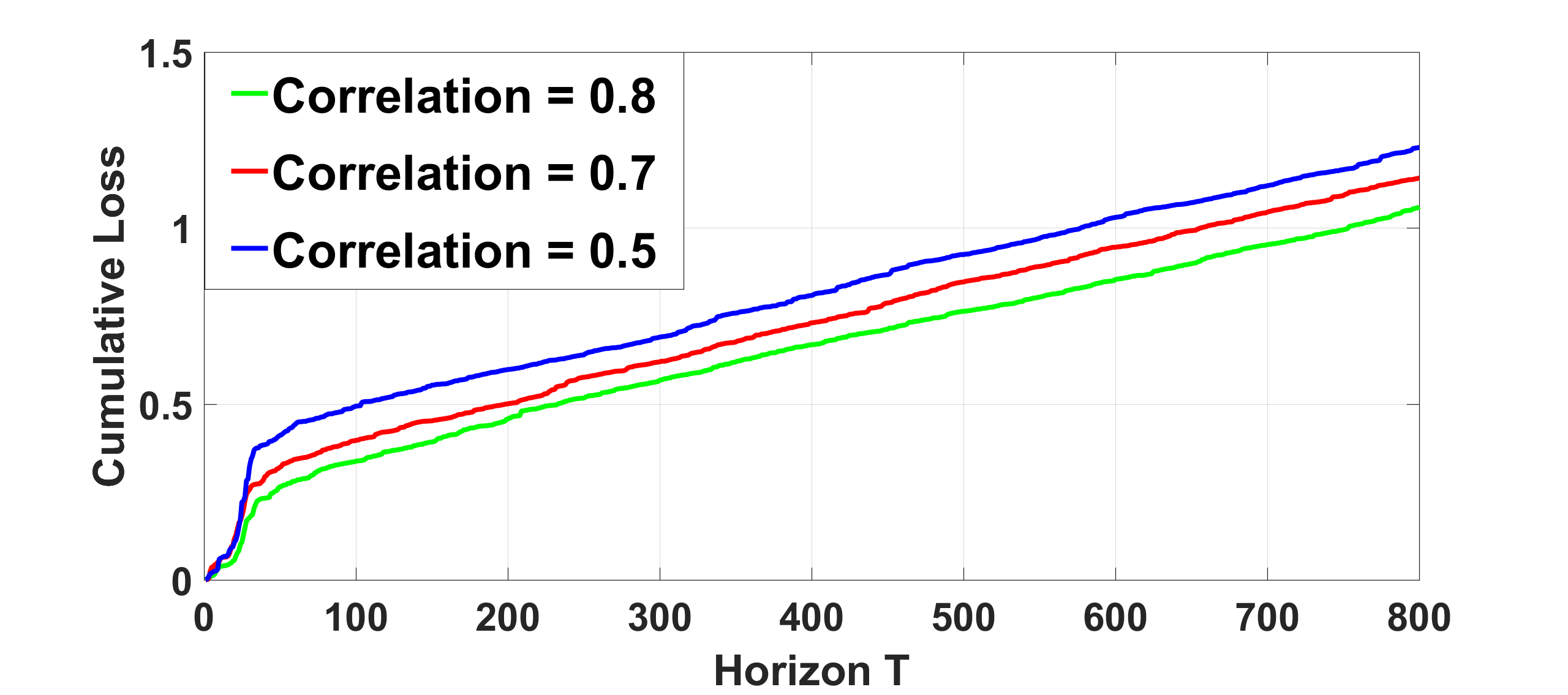}
    \vspace{-1mm}
    \caption{The evolution of cumulative loss with respect to time for different correlation coefficients between $x_u^t$ and $x_k^t$ . Each point in the plot is an average over $100$ trials.
     }
    \label{fig:correlation}
\end{figure}

\vspace{2mm}
Now, keeping everything fixed, we generate $x_u^t$ and $x_k^t$ from a correlated gaussian density with varying correlation factor and we would like to see how  correlation plays a role in estimating $x_u^t$. We fix the delay $\tau =10$. Figure~\ref{fig:correlation} shows the behavior of cumulative loss for different correlation factors. It is intuitive to argue that the correlation function plays a better role when $x_u^t$ and $x_k^t$ are highly correlated. This is observed in Figure~\ref{fig:correlation}.

\subsection{Comparison with a Naive Heuristic algorithm}

Now we compare the delayed gradient approach of estimating $x_u^t$ to a naive approach based on sample averaging. We work with fixed delay setting and assume there is no correlation between $x_u^t$ and $x_k^t$ (i.e., $\Phi(.)=0$). At time $t+1$, since the information upto time $t-\tau$ (in this case $x_u^k$ for $k=1,\ldots,t-\tau$) is known, one can simply form an estimate, $\tilde{x}_u^{t+1}= \frac{1}{t-\tau} \sum_{k=1}^{t-\tau} x_u^k$. If $x_u^t$ is drawn from a well behaved distribution (e.g. sub-gaussian, sub-exponential) and the delay is small with respect to the time horizon $T$, we can expect the naive estimator to work well since empirical average is the minimum variance unbiased estimator of the mean of the distribution and the concentration of measure phenomenon ensures that the samples are close to the mean with high probability. However, when $x_u^t$ is drawn from some arbitrary convex set, there is no guarantee on the performance of the naive estimator. Since our framework is general enough to handle samples from arbitrary convex set, the guarantees on cumulative loss function continue to hold.

\begin{figure}[t!]
    \centering
    \includegraphics[height = 0.25\textwidth, width=0.4\textwidth]{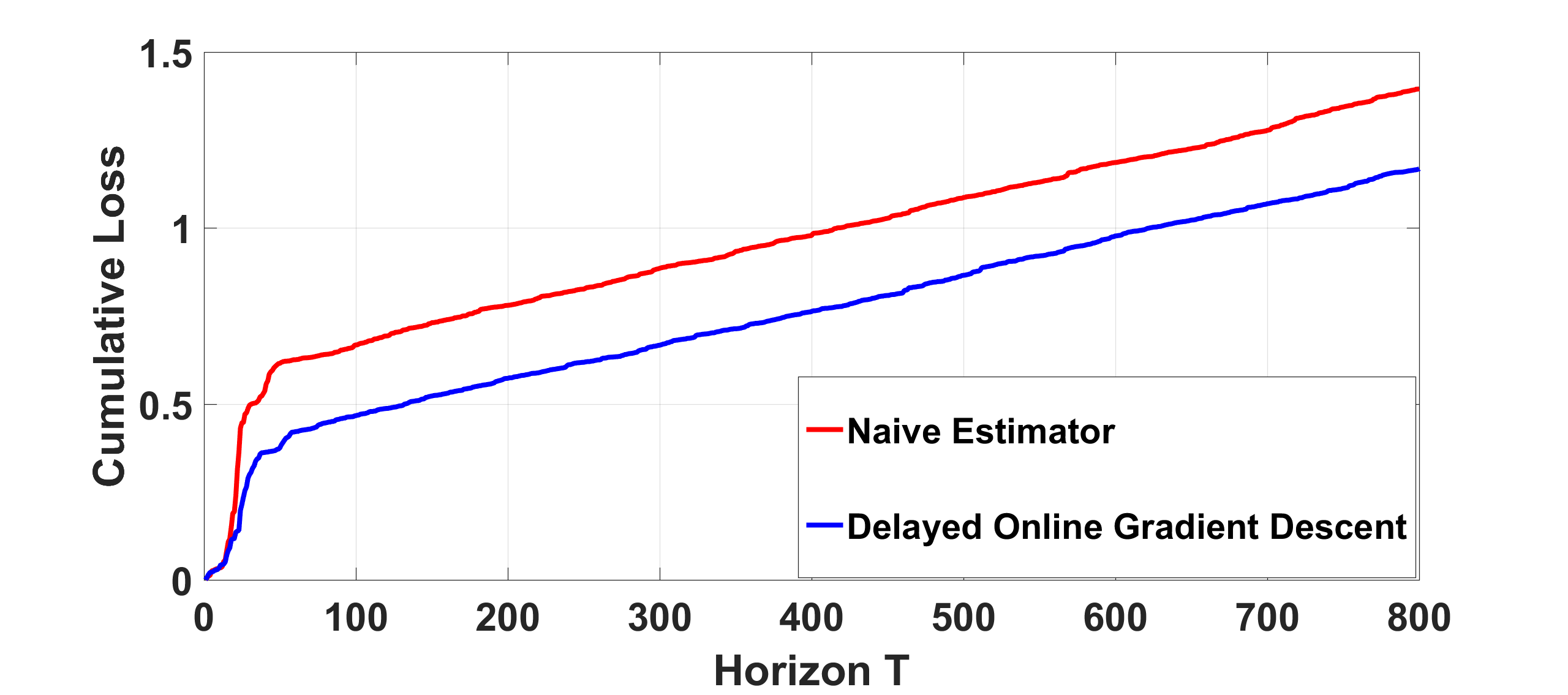}
    \vspace{-1mm}
    \caption{Cumulative loss with respect to time horizon for samples ($x_u^t$) drawn from normal distribution. Each point in the plot is an average over $100$ trials.
     }
    \label{fig:normal_draw}
\end{figure}

\begin{figure}[t!]
    \centering
    \includegraphics[height = 0.25\textwidth, width=0.4\textwidth]{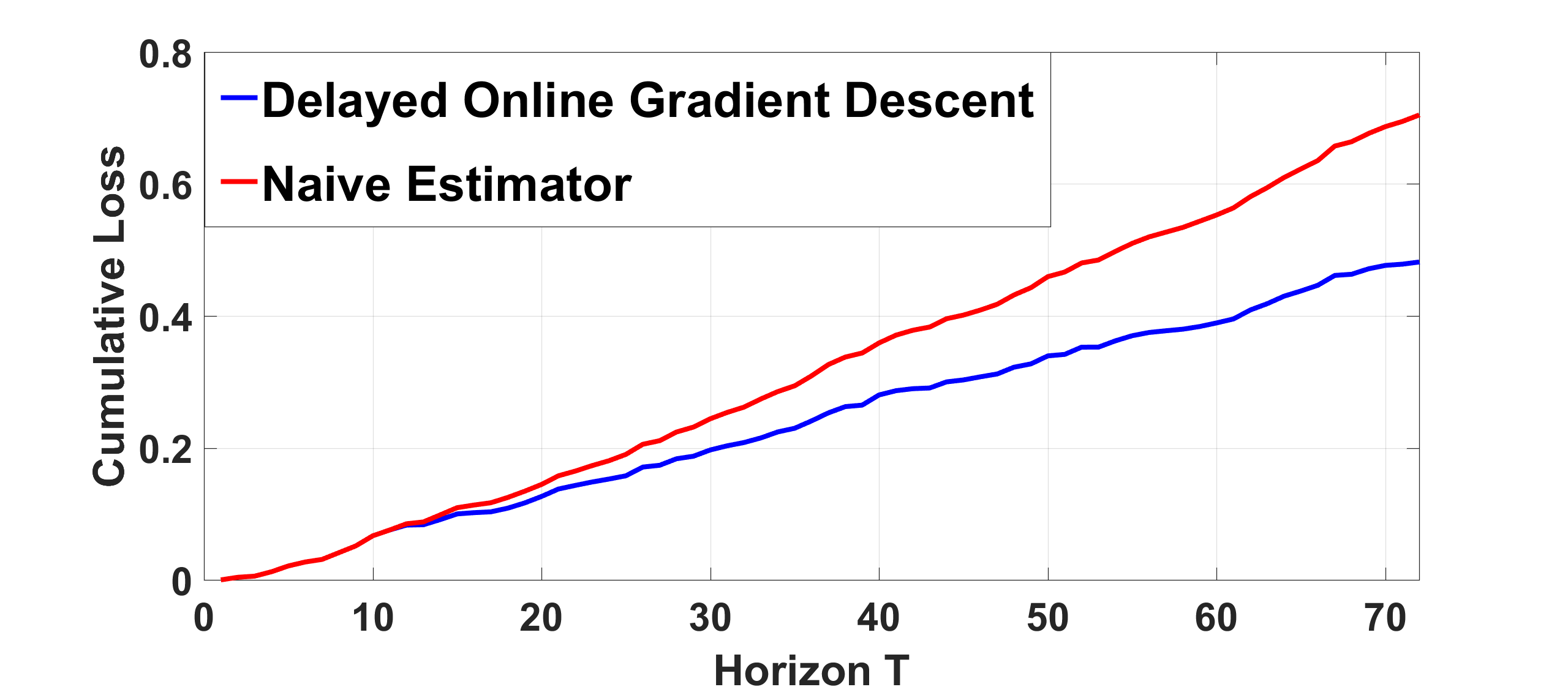}
    \vspace{-1mm}
    \caption{Cumulative loss with respect to time when $x_u^t$ is drawn from a pentagon. Each point in the plot is an average over $100$ trials.
     }
    \label{fig:pentagon}
\end{figure}

We now simulate two different setting to demonstrate the performance of the delayed online gradient estimator and the naive estimator: a) when $x_u^t$ is drawn from a normal distribution and b) when $x_u^t$ is drawn from a pentagon (which is a convex set). The performance is shown in Figure~\ref{fig:normal_draw} and \ref{fig:pentagon}. Since normal distributions are sub-gaussian, owing to concentration of measure phenomena, we observe that the naive estimator performs reasonably well, but for the second setting (samples drawn from a pentagon), the naive algorithm performs poorly. This validates the strength and robustness (i.e., it is independent of the distribution) of the online delayed gradient descent algorithm.

\section{Conclusion and Future Work}
We consider an online optimization approach to tackle the problem of inference with partial information. Although we can output a score estimate with this approach, we cannot predict the absolute (which also includes the candidates not seen upto now) ranking of the agents. Our immediate future direction is to tackle the issue of online absolute ranking. Also, we would like to extend the current formulation to a) a setting with bandit feedback, i.e., the loss function will not be known, only the value of the loss function at the chosen action will be revealed and b) a setting where the loss function is possibly non-convex. We keep these as our future endeavors.

\bibliographystyle{IEEEtran}
\bibliography{IEEEabrv,ref}

\section*{APPENDIX}
\section{Online Learning with Fixed Delay}
\label{appendix:fixed_delay}

\subsection{Proof of Lemma~\ref{lem:divergence}}

From the definition of $D(.||.)$, we have,
\begin{eqnarray}
&& \hspace{-6mm} D(x^*||\hat{x}^{t+1}_u)-D(x^*|| \x) =\frac{1}{2}\norms{x^* - \hat{x}_u^{t+1}}^2-\frac{1}{2}\norms{x^*  \nonumber \\ 
&-& \x}^2 = \frac{1}{2}\norms{x^* - \x + \eta_t \tilde{g}_{t-\tau}}^2-\frac{1}{2}\norms{x^* - \x}^2 \nonumber \\
&=& \frac{1}{2}\eta^2 \norms{\tilde{g}_{t-\tau}}^2 - \eta_t \langle \x-x^*,\tilde{g}_{t-\tau} \rangle = \frac{1}{2}\eta^2 \norms{\tilde{g}_{t-\tau}}^2 \nonumber \\
&-& \eta_t \langle \hat{x}_u^{t-\tau} -x^*,\tilde{g}_{t-\tau} \rangle + \eta_t \langle \hat{x}_u^{t-\tau}- \x , \tilde{g}_{t-\tau} \rangle \nonumber
\end{eqnarray}

We now can further simplify the last term as follows: since after the initialization phase, $ \tau \leq t < 2\tau$, we obtain $t-\tau$ gradients, we have,
\begin{eqnarray}
\langle \x -\hat{x}_u^{t-\tau}, \tilde{g}_{t-\tau} \rangle = \hspace{-3mm} \sum_{k=1}^{\min(\tau,t-(\tau+1))} \hspace{-3mm}\langle \hat{x}_u^{t-(k-1)}-\hat{x}_u^{t-k}, \tilde{g}_{t-\tau} \rangle \nonumber
\end{eqnarray}
 
Now the proof follows by plugging $\hat{x}_u^{t-(k-1)}$ and rearranging the terms.

\subsection{Proof of Theorem~\ref{thm:convex_fixed}}

Summing over Lemma~\ref{lem:divergence}, we get
\begin{eqnarray}
&&\sum_{t=\tau+1}^{T+\tau} \langle \hat{x}^{t-\tau}_u -x^*, \tilde{g}_{t-\tau} \rangle \nonumber \\
&& = \sum_{t=\tau+1}^{T+\tau}  ( \frac{1}{2}\eta_t \norms{\tilde{g}_{t-\tau}}^2 + \hspace{-5mm} \sum_{k=1}^{\min(\tau,t-(\tau+1))} \hspace{-4mm} \eta_{t-k} \langle \tilde{g}_{t-\tau-k},  \tilde{g}_{t-\tau} \rangle  ) \nonumber \\
&+& \frac{D(x^*||\hat{x}_u^{\tau+1})}{\eta_{\tau+1}}-\frac{D(x^*||\hat{x}_u^{T+\tau+1})}{\eta_{T+\tau}} \nonumber \\
&+&\sum_{t=\tau+2}^{T+\tau} \bigg ( D(x^*||\x)[\frac{1}{\eta_t}-\frac{1}{\eta_{t-1}}] \bigg ) \label{eqn:reg_inner_prod}
\end{eqnarray}

We will now separately control the terms in the above equation. Via the Lipschitz property of $\tilde{g}_{t-\tau}$, we have

\begin{eqnarray}
\sum_{t=1+\tau}^{T+\tau} \frac{1}{2}\eta_t \norms{\tilde{g}_{t-\tau}}^2 \leq \frac{1}{2}L'^2 \sum_{t=\tau+1}^{T+\tau} \eta_t \leq \sigma L'^2 \sqrt{T} \nonumber
\end{eqnarray}

where the last inequality is derived from the fact that, $\sum_{k=1}^{T}\frac{1}{2\sqrt{k}} \leq \int_{0}^{T}\frac{1}{2\sqrt{x}}dx = \sqrt{T}$.

Since, $\frac{D(x^*||\hat{x}_u^{T+\tau+1})}{\eta_{T+\tau}}$ is positive, we can get rid of it while upper-bounding Equation~\ref{eqn:reg_inner_prod}. Using the fact that, $D(x_1||x_2) \leq 2R^2$, we have,

\begin{eqnarray}
&& \frac{D(x^*||\hat{x}_u^{\tau+1})}{\eta_{\tau+1}}+\sum_{t=\tau+2}^{T+\tau} \bigg ( D(x^*||\x)[\frac{1}{\eta_t}-\frac{1}{\eta_{t-1}}] \bigg ) \nonumber \\
 && \leq \frac{2R^2}{\sigma}+\frac{2R^2}{\sigma}\sum_{t=\tau+2}^{T+\tau} (\sqrt{t-\tau}-\sqrt{t-\tau -1}) \nonumber \\
 && = \frac{2R^2}{\sigma}\sqrt{T} \nonumber
\end{eqnarray}

We now analyze the term involving $\min(\tau,t-(\tau+1))$. The dependence on $\tau$ comes from this term. 
\begin{eqnarray}
&&\sum_{k=1}^{\min(\tau,t-(\tau+1))} \eta_{t-k} \langle \tilde{g}_{t-\tau-k},\tilde{g}_{t-\tau} \rangle \nonumber \\
&&  \leq  \sum_{k=1}^{\min(\tau,t-(\tau+1))} \eta_{t-k} (L+\frac{\sigma R}{\sqrt{t-\tau-k}})\norms{\tilde{g}_{t-\tau}} \nonumber \\
& \leq & \min(\tau,t-(\tau+1)) \eta_{t-\min(\tau,t-(\tau+1))} \nonumber \\
&& \times (L+\frac{\sigma R}{\sqrt{t-\tau - \min(\tau,t-(\tau+1))}})\norms{\tilde{g}_{t-\tau}} \nonumber
\end{eqnarray}

Now we have,
\begin{eqnarray}
&&\sum_{t=\tau+1}^{T+\tau} \min(\tau,t-(\tau+1)) \eta_{t-\min(\tau,t-(\tau+1))} \nonumber \\
&& \times (L+\frac{\sigma R}{\sqrt{t-\tau - \min(\tau,t-(\tau+1))}})\norms{\tilde{g}_{t-\tau}} \nonumber\\
&&\leq \sum_{t=\tau+1}^{2\tau} (t-\tau-1) \eta_{\tau+1}(L+\frac{\sigma R}{\sqrt{1}})L' +\sum_{t=2\tau+1}^{T+\tau} \tau \eta_{t-\tau} (L \nonumber \\
&& +\frac{\sigma R}{\sqrt{t-2\tau}})L' \leq \sum_{t=\tau+1}^{2\tau} \frac{\sigma}{\sqrt{1}}L'^2 + \sum_{t=2\tau+1}^{T+\tau} \tau \frac{\sigma}{\sqrt{t-2\tau}} L'^2 \nonumber \\
&& \leq \frac{\sigma \tau^2}{2} L'^2+ 2\sigma\tau \sqrt{T}L'^2 \nonumber
\end{eqnarray}

Now, putting everything together, the regret is computed as,
\small
\begin{eqnarray}
&& R(T)\leq  \sum_{t=\tau+1}^{T+\tau} \langle \hat{x}_u^{t-\tau}-x^* \rangle + \sum_{t=\tau+1}^{T+\tau} \langle \Phi_{t+1}(x_k^{t+1}), \hat{x}_u^{t-\tau}-x^* \rangle \nonumber \\
&& \leq \sigma L'^2 \sqrt{T} + \frac{2R^2}{\sigma}\sqrt{T} + \frac{\sigma \tau^2}{2} L'^2+ 2\sigma\tau \sqrt{T}L'^2 + 2R^2 \hspace{-2mm} \sum_{t=\tau+1}^{T+\tau}  \hspace{-3mm}\lambda_{t+1} \nonumber \\
&& \leq \bigg (\sigma L'^2 + \frac{2R^2}{\sigma} + 2\sigma \tau L'^2 + 4R^2 \sigma \bigg ) \sqrt{T} + \frac{\sigma \tau^2}{2}L'^2  \nonumber
\end{eqnarray}

Now, plugging the value of $\sigma (:=\frac{R}{L'\sqrt{\tau}})$, we have,
\begin{equation}
R(T)=\mathcal{O}(\sqrt{\tau T}) \nonumber
\end{equation}

\subsection{Proof of Theorem~\ref{thm:strongly_convex_fixed}}
From the strong convexity of the loss functions,
\begin{eqnarray}
&& R(T)  \leq  \sum_{t=\tau +1 }^{T+\tau}  \bigg [ \langle \hat{x}_u^{t-\tau} - x^*, \tilde{g}_{t-\tau} \rangle + \langle \frac{\beta_t}{\eta_t} \phi_{t+1}(x_k^{t+1}), \nonumber \\
&& \x - x^* \rangle  -\frac{\gamma}{2}\norms{\hat{x}_u^{t-\tau} - x^*}^2  \bigg ] 
 \leq  \sum_{t=\tau +1 }^{T+\tau} \bigg [ \frac{1}{2}\eta_t \norms{\tilde{g}_{t-\tau}}^2 \nonumber \\
 && +\frac{D(x^*||\x)-D(x^*||\hat{x}^{t+1}_u)}{\eta_t}+ \hspace{-6mm} \sum_{k=1}^{\min(\tau,t-(\tau+1))} \hspace{-6mm} \eta_{t-k} \langle \tilde{g}_{t-\tau-k},  \nonumber \\
  && \tilde{g}_{t-\tau} \rangle + \langle \frac{\beta_t}{\eta_t} \phi_{t+1}(x_k^{t+1}), \x -x^* \rangle   -\frac{\gamma}{2}\norms{\hat{x}_u^{t-\tau} - x^*}^2  \bigg ] \nonumber \\
&& \leq \sum_{t=\tau +1 }^{T+\tau} \bigg [ (\frac{\eta_t}{2} + \tau \eta_{\max(\tau,t-\tau-1)}) L'^2 + \gamma (t-\tau) \nonumber \\
&& \times (D(x^*||\x)-D(x^*||\hat{x}^{t+1}_u) )
 -\gamma D(x^*||\hat{x}_u^{t-\tau}) \nonumber \\
 && + \langle \frac{\beta_t}{\eta_t} \phi_{t+1}(x_k^{t+1}), \x -x^* \rangle   -\frac{\gamma}{2}\norms{\hat{x}_u^{t-\tau} - x^*}^2  \bigg ] \nonumber 
\end{eqnarray}

Now, we can see a few component telescopes,

\begin{eqnarray}
&& R(T) \leq \sum_{t=\tau +1 }^{T+\tau} \bigg [ \frac{\eta_t}{2} + \tau \eta_{\max(\tau,t-\tau-1)} \bigg ] L'^2 \nonumber \\
&&  + \frac{1}{\gamma} \sum_{t=\tau+1}^{T+\tau} \frac{2 R^2  }{ t-\tau } + \sum_{t=1}^{\tau} \gamma (D(x^*||x_{T+1})-D(x^*||x_t)) \nonumber \\
 &&  \leq  2\gamma \tau R^2 + \frac{2R^2 H(T)}{\gamma}  + \hspace{-3mm} \sum_{t=\tau +1 }^{T+\tau} \hspace{-1.5mm} [ \frac{\eta_t}{2} + \tau \eta_{\max(\tau,t-\tau-1)}] L'^2 \nonumber
\end{eqnarray}

where $H(T)$ is the $T-$th harmonic number.  Since $\eta_t$ is monotonically decreasing with $t$ (for $t \geq \tau +1 $), we have, $\eta_t \leq \eta_{\max(\tau,t-\tau-1)}$ for $t \geq \tau +1 $. Then, we have,

\small
\begin{eqnarray}
 R(T) \leq  2\gamma \tau R^2+ \frac{2R^2}{\gamma} H(T) + (\frac{1}{2}+\tau) L'^2 \sum_{t=\tau +1}^{T+\tau} \hspace{-3mm}  \eta_{\max(\tau,t-\tau-1)} \nonumber \\
 \leq 2\gamma \tau R^2+ \frac{2R^2}{\gamma} H(T) + (\frac{1}{2}+\tau) L'^2 \sum_{t=1}^{T}  \eta_{\max(\tau,t-1)} \nonumber \\
 =  2\gamma \tau R^2+ \frac{2R^2}{\gamma} H(T) + (\frac{1}{2}+\tau) L'^2 \sum_{t=\tau +1 }^{T}  \eta_{t} \nonumber \\
 = 2\gamma \tau R^2+ \frac{2R^2}{\gamma} H(T) + (\frac{1}{2}+\tau) L'^2 \frac{1}{\gamma} H(T-\tau) \nonumber 
\end{eqnarray}

when $T$ is reasonably large, $H(T) \approx \log T$ and $H(T-\tau) \approx \log (T-\tau)$. Plugging the values will yield the result.

\subsection{Proof of Theorem~\ref{thm:mirror_fixed}}
We can write,
\small
$$
\langle \x-x^*,\tilde{g}_{t-\tau} \rangle = \langle \hat{x}_u^{t-\tau}-x^*,\tilde{g}_{t-\tau} \rangle + \sum_{k=0}^{\tau -1} \langle \hat{x}_u^{t-k} - \hat{x}_u^{t-k-1}, \tilde{g}_{t-\tau} \rangle
$$
We use the assumption on the mirror map to the summation in the above equation, we have,
 $$
\langle \x-x^*,\tilde{g}_{t-\tau} \rangle \geq \langle \hat{x}_u^{t-\tau}-x^*,\tilde{g}_{t-\tau} \rangle - \tau \eta_{t-\tau} L_M L'^2
$$
Now using Lemma~\ref{lem:mirror}, and summing from $\tau+1$ to $T+\tau$, using the same machinery as of Theorem~\ref{thm:convex_fixed}, we obtain the given regret bound.

\normalsize
\subsection{Proof of Theorem~\ref{thm:convex_adversary}}
We proceed by splitting the sum over gradients, and analyzing it separately. We borrow a few techniques from \cite{quanrud_15}. Let  $s \in \mathcal{F}_t$, and $\mathcal{F}_{t,s} = \lbrace t_1 \in \mathcal{F}_t: s>t_1 \rbrace$. Let
$$
\hat{x}_u^{t,s} = \x -\eta  \hspace{-3mm}\sum_{t' \in \mathcal{F}_{t,s}} \hspace{-3mm} g_{t'} +  \beta \Phi_t(x_k^{t+1}) = \x - \eta  \hspace{-3mm} \sum_{t' \in \mathcal{F}_{t,s}} \hspace{-3mm} g_{t'} + \eta \lambda x_k^{t+1}
$$
Also let $s'=\max \mathcal{F}_t$. Therefore $s'$ denotes the latest available information at time $t$. We have,

\begin{eqnarray}
&& \norms{\hat{x}_u^{t+1} - x^*}^2 = \norms{ \hat{x}_u^{t,s'}-\eta g_{s'} -x^*}^2 \nonumber \\
&& = \norms{\hat{x}_u^{t,s'}-x^*}^2-2 \eta \langle g_{s'},\hat{x}_u^{t,s'}-x^* \rangle +\eta^2 \norms{g_{s'}}^2 \nonumber
\end{eqnarray}

Continue unrolling the first term, we get,
\begin{eqnarray}
\norms{\hat{x}_u^{t+1} - x^*}^2 &=& \norms{\x -x^*}^2 + 2\eta \sum_{s\in \mathcal{F}_t} \langle g_s, x^* - \hat{x}_u^{t,s} \rangle \nonumber \\
& + & \eta^2 \sum_{s \in \mathcal{F}_t} \norms{g_s}^2 \nonumber
\end{eqnarray}

Invoking the convexity of $f$, we have,
\begin{eqnarray}
&& \langle g_s, x^* - \hat{x}_u^{t,s} \rangle = \langle g_s, x^* - \hat{x}_u^s \rangle +  \langle g_s, \hat{x}_u^s - \hat{x}_u^{t,s} \rangle \nonumber \\
&& \leq f_s(x^*) - f_s (\hat{x}_u^s) + \langle g_s, \hat{x}_u^s - \hat{x}_u^{t,s} \rangle \nonumber
\end{eqnarray}

With the assumption $\norms{g_t} \leq L$ for all $t$, the regret,
\begin{eqnarray}
&& R(T)= \sum_{t=1}^{T} f_t(\x)-f_t(x^*)=\sum_{t=1}^{T} \sum_{s \in \mathcal{F}_t} f_s (\hat{x}_u^s)-f_s(x^*) \nonumber \\
&& \leq \frac{1}{2\eta}\sum_{t=1}^{T} \bigg (\norms{\x -x^*}^2 -\norms{\hat{x}_u^{t+1}-x^*}^2 + 2\eta \sum_{s \in \mathcal{F}_t} \langle g_s, \hat{x}_u^s - \hat{x}_u^{s,t} \rangle \nonumber \\
&& + \eta^2 |\mathcal{F}_t| L^2 \bigg )  = \frac{1}{2\eta}\sum_{t=1}^{T}  (\norms{\x -x^*}^2 -\norms{\hat{x}_u^{t+1}-x^*}^2  ) +\frac{\eta}{2}L^2 T \nonumber \\
&&  + \sum_{t=1}^{T} \sum_{s \in \mathcal{F}_t} \langle g_s, \hat{x}_u^s - \hat{x}_u^{s,t} \rangle
 \leq \frac{1}{2\eta} +\frac{\eta}{2}L^2 T + \sum_{t=1}^{T} \sum_{s \in \mathcal{F}_t} \langle g_s, \hat{x}_u^s - \hat{x}_u^{s,t} \rangle \nonumber
\end{eqnarray}

where the last inequality follows from telescoping. 
\begin{eqnarray}
&&\sum_{t=1}^{T} \sum_{s \in \mathcal{F}_t} \langle g_s, \hat{x}_u^s - \hat{x}_u^{s,t} \rangle \leq \sum_{t=1}^{T} \sum_{s \in \mathcal{F}_t} L \norms{\hat{x}_u^s - \hat{x}_u^{s,t}} \nonumber \\
&& \leq L \sum_{t=1}^{T} \sum_{s \in \mathcal{F}_t} \norms{ \hat{x}_u^s -  \x + \eta  ( \sum_{t' \in \mathcal{F}_{t,s}} g_{t'})} + \eta \lambda L \sum_{t=1}^{T} \sum_{s \in \mathcal{F}_t} \norm{ x_k^{t+1}}  \nonumber 
\end{eqnarray}
We now invoke the following result proved in [\cite{quanrud_15}, Theorem 2.1]:
$$
 L \sum_{t=1}^{T} \sum_{s \in \mathcal{F}_t} \norms{ \hat{x}_u^s -  \x + \eta  ( \sum_{t' \in \mathcal{F}_{t,s}} g_{t'})} \leq 2\eta L^2 \sum_{t-1}^{T}d_t = 2\eta L^2 D
$$
Therefore,
\begin{eqnarray}
&& \sum_{t=1}^{T} \sum_{s \in \mathcal{F}_t} \langle g_s, \hat{x}_u^s - \hat{x}_u^{s,t} \rangle \leq 2\eta L^2 D + \eta |\lambda| L R \sum_{t=1}^{T} |\mathcal{F}_t| \nonumber \\
&& = \eta (2L^2 D + |\lambda| L R T) \nonumber
\end{eqnarray}

 Plugging in,

Putting everything together,
$$
R(T) \leq \frac{1}{2\eta}+ \eta (\frac{TL^2}{2}+ |\lambda| L R T + 2L^2 D)
$$
Now, choose $\eta$ such that,
$$
\frac{1}{\eta^2} = T(L^2 + 2 |\lambda| L R)+ 4 L^2 D \Rightarrow \eta = \mathcal{O}(1/\sqrt{T+D})
$$
Hence,
$$
R(T)= \mathcal{O}(\sqrt{D+T})=\mathcal{O}(\sqrt{D})
$$

\end{document}

%% file: newcommands.tex
\usepackage{hyperref}

\newcommand{\R}{\mathbb{R}}

\newcommand{\norm}[1]{\left\|#1\right\|}
\newcommand{\norms}[1]{\|#1\|}

\newcommand{\twonms}[1]{\|#1\|_2}
